\documentclass{article}

\PassOptionsToPackage{numbers, sort}{natbib}

\usepackage[preprint]{neurips_2021_ml4ps}

\usepackage[utf8]{inputenc} %
\usepackage[T1]{fontenc}    %
\usepackage{hyperref}       %
\usepackage{url}            %
\usepackage{booktabs}       %
\usepackage{amsfonts}       %
\usepackage{nicefrac}       %
\usepackage{microtype}      %
\usepackage{xcolor}         %

\usepackage{graphicx} %
\usepackage{caption} %
\usepackage{subcaption} %
\usepackage{amsmath} %

\DeclareMathOperator{\enc}{enc}
\DeclareMathOperator{\dec}{dec}

\DeclareMathOperator{\trend}{trend}
\DeclareMathOperator{\scale}{scale}
\DeclareMathOperator{\shift}{shift}
\DeclareMathOperator{\cut}{cut}
\DeclareMathOperator{\resize}{resize}
\DeclareMathOperator{\interpl}{interpl}

\newcommand{\id}{\boldsymbol{I}}

\newcommand{\bx}{\boldsymbol{x}}
\newcommand{\by}{\boldsymbol{y}}
\newcommand{\bz}{\boldsymbol{z}}
\newcommand{\bt}{\boldsymbol{t}}

\newcommand{\bxi}{\boldsymbol{\xi}}
\newcommand{\btheta}{\boldsymbol{\theta}}
\newcommand{\bphi}{\boldsymbol{\phi}}
\newcommand{\bmu}{\boldsymbol{\mu}}
\newcommand{\bsigma}{\boldsymbol{\sigma}}

\newcommand{\bxhat}{\hat{\boldsymbol{x}}}
\newcommand{\byhat}{\hat{\boldsymbol{y}}}
\newcommand{\bthat}{\hat{\boldsymbol{t}}}
\newcommand{\bxihat}{\hat{\boldsymbol{\xi}}}

\title{Automatically detecting anomalous exoplanet transits}

\author{%
    Christoph J. H\"ones \\
    University of Amsterdam \\
    \texttt{christoph.hoenes@googlemail.com} \\
    \And
    Benjamin Kurt Miller \\
    University of Amsterdam \\
    \texttt{b.k.miller@uva.nl} \\
    \And
    Ana M. Heras \\
    European Space Agency, ESTEC \\
    \texttt{ana.heras@esa.int} \\
    \And
    Bernard Foing \\
    Leiden Observatory, VU Amsterdam, \\ ISU, ILEWG EMMESI \\
}

\begin{document}

\maketitle

\begin{abstract}
    Raw light curve data from exoplanet transits is too complex to naively apply traditional outlier detection methods. 
    We propose an architecture which estimates a latent representation of both the main transit and residual deviations with a pair of variational autoencoders. 
    We show, using two fabricated datasets, that our latent representations of anomalous transit residuals are significantly more amenable to outlier detection than raw data or the latent representation of a traditional variational autoencoder.
    We then apply our method to real exoplanet transit data.
    Our study is the first which automatically identifies anomalous exoplanet transit light curves.
    We additionally release three first-of-their-kind datasets to enable further research.
\end{abstract}

\section{Introduction}
\label{sec:introduction}

\paragraph{Exoplanets}
An exoplanet obscuring the line-of-sight between a star and telescope, thereby temporarily reducing the observed brightness (light curve) of that star, is known as a transit \cite{perryman2018exoplanet_handbook}. Today, detected transits number in the thousands \cite{nasa2021exoplanet_archive} and are normally identified using variants of the Box Least Squares algorithm \cite{kovacs2002bls, hippke2019tls}. 
Deviation from a theoretical transit light curve model \cite{mandel2002analytic} indicates unusual or unknown astrophysical phenomena (e.g. \cite{silvavalio2010corot_spots, ahlers2020gd_analysis, lieshout2018dre}). Therefore, finding anomalous transit light curves is of great scientific interest. 
Large-scale space telescope surveys and improvements in automated detection methods will lead to unprecedentedly high detection rates of transits, demanding a fast outlier detection solution. We propose to automate the identification of anomalous transits. 

\begin{figure}[hbt]
    \centering
    \begin{subfigure}[b]{0.9\textwidth}
         \centering
         \includegraphics[width=\textwidth]{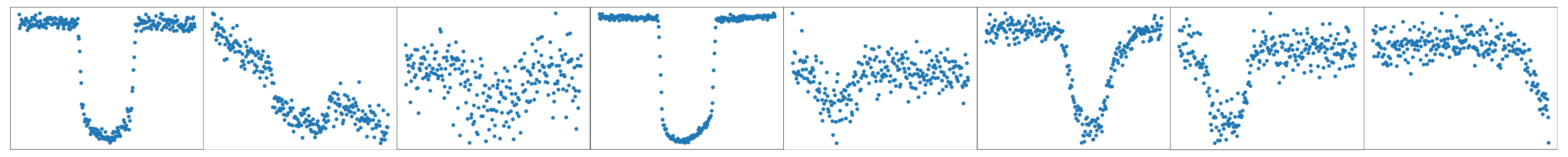}
         \caption{Real TESS data}
         \label{fig:real_transit_examples}
    \end{subfigure}
    \begin{subfigure}[b]{0.45\textwidth}
         \centering
         \includegraphics[width=\textwidth]{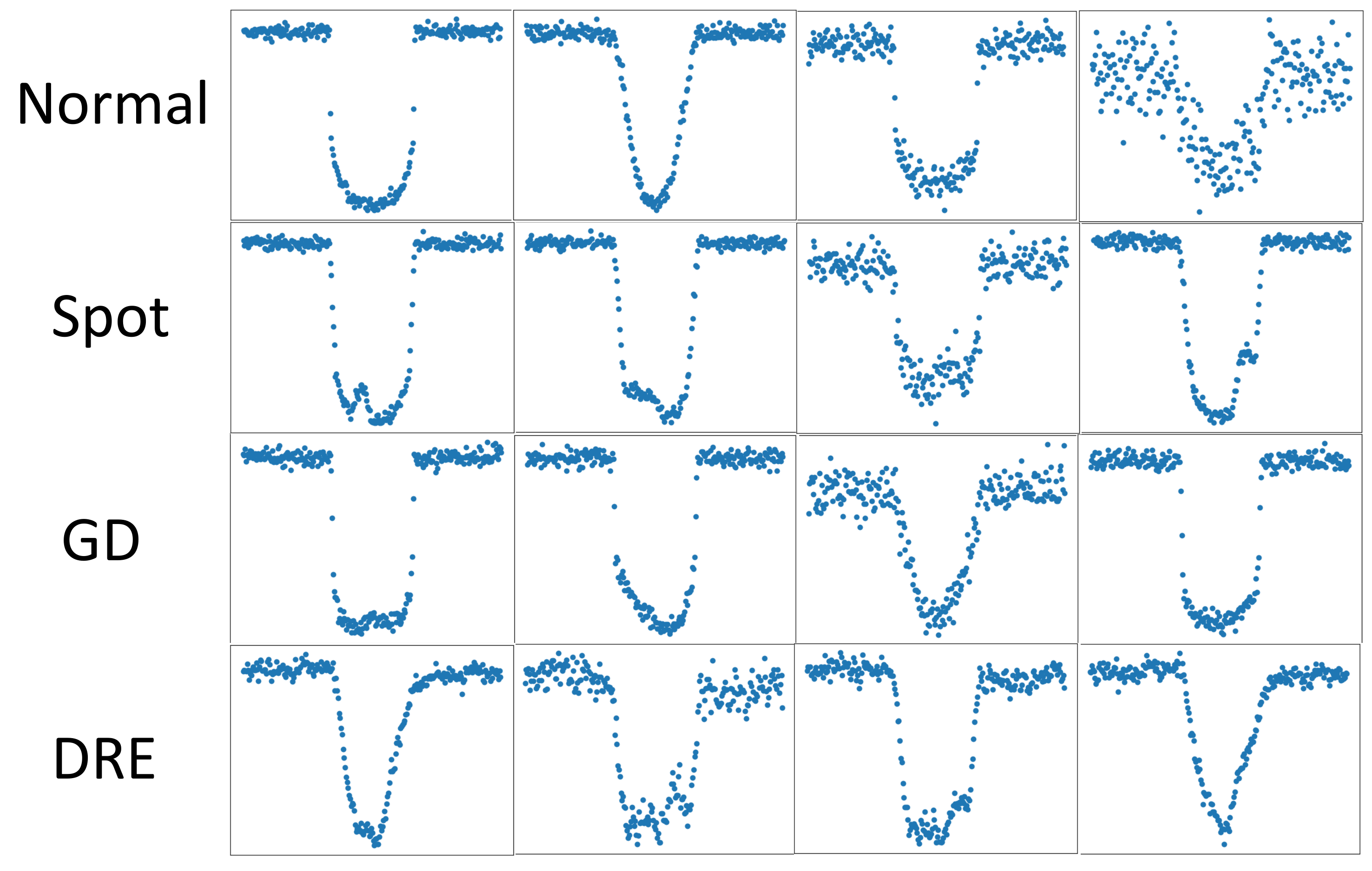}
         \caption{ALT-i}
         \label{fig:alt_i_examples}
    \end{subfigure}
    \begin{subfigure}[b]{0.45\textwidth}
         \centering
         \includegraphics[width=\textwidth]{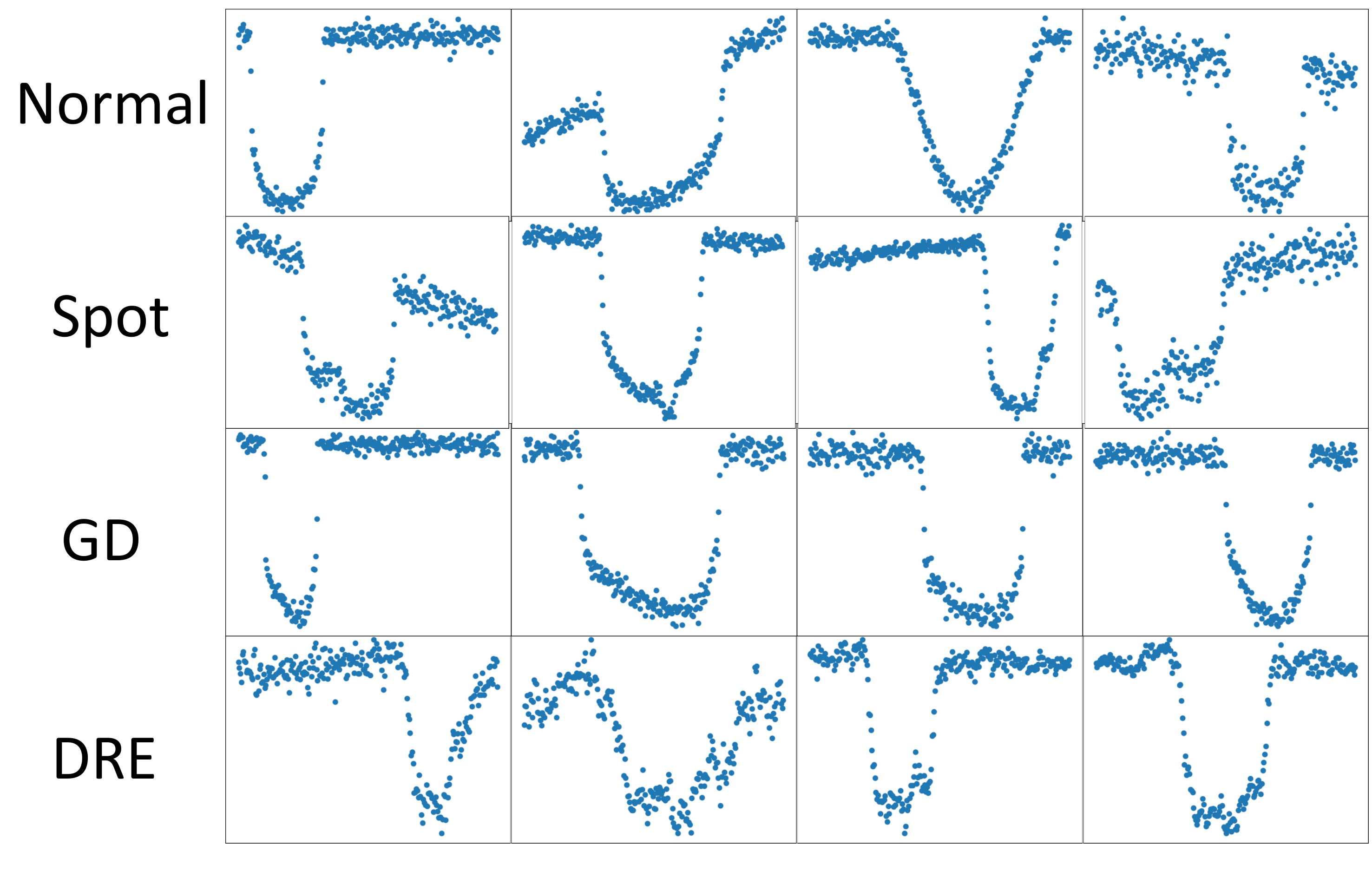}
         \caption{ALT-h}
         \label{fig:alt_h_examples}
    \end{subfigure}
  \caption{Examples of fabricated and genuine transit light curves. Genuine TESS data on top. Fabricated data from ALT-i and ALT-h on the bottom. The y-axes represent the normalized (unitless) flux (brightness of the star) and the x-axes represent the phase (time relative to the transit duration).}
  \label{fig:transit_examples}
\end{figure}

\paragraph{Related work} 
Deep learning has been applied to exoplanet detection \cite{pearson2018_neural_detection} and confirmation, also known as automated vetting \cite{mccauliff2015autovetter, thompson2015robovetter, shallue2018astronet}. Feature extraction and clustering have been applied to Kepler data \cite{thompson2016kepler, kepler} to find stellar variability outliers \cite{giles2018ml_kepler_anomaly}. Similarly, 
\cite{nun2016ensemble_lc_anomaly} use an ensemble of outlier detection methods on features extracted from light curve data to look for special stellar activity. 

\paragraph{Contribution} This paper is concerned with an entirely novel topic: Automatically identifying anomalous exoplanet transit light curves. (a) Our paper represents the first systematic study of automated detection of exoplanet transit shape anomalies. (b) Our proposed architecture assigns an anomaly score to each transit light curve. On one fabricated dataset with known labels, our architecture produces an average precision 61.81\% higher than using anomaly detection on raw data alone and 23.08\% higher than a naive variational autoencoder-based dimension reduction technique \cite{kingma2013auto, rezende2014stochastic}. (c) We present two sets of fabricated, labeled transit data along with an unlabeled collection of genuine transit data.
\footnote{
    The source code and the data can be found at: \\ \url{https://github.com/ChristophHoenes/AnomalousExoplanetTransits}
}
Each dataset is the first-of-its-kind for deep learning methods, thereby catalyzing future deep learning research into exoplanetary science.

\section{Genuine and fabricated transit light curves}
\label{sec:data}

\paragraph{Genuine transit data} 
We curated a set of transit light curves with a total of 1668 targets from the Transiting Exoplanet Survey Satellite (TESS) mission \cite{ricker2014tess}, which is publicly available at MAST \cite{mast}, using the \texttt{astroquery} \cite{ginsburg2019astroquery} package.
As the data is unlabeled, we can make no guarantees about the distribution of anomalies to regular transit light curves. Data processing closely follows the pipelines of exoplanetary scientists. The most important processing step is phase-folding where the time series is transformed into phase space (time relative to the orbital period). If the parameters used for phase-folding are accurate, the transits within a light curve perfectly overlay and have zero phase at the transit mid-point.
Inaccuracies in the currently available parameter estimates of period, epoch and transit duration can lead to \textit{horizontal shifts} of the transit shape within the extracted observation window or different \textit{horizontal scale} (e.g. apparent transit width within the observation window). This needs to be considered in the model design. Further details can be found in Appendix \ref{appendix:data}.

\paragraph{Fabricated transit data}
Quantification of the precision and recall of anomaly detection requires labeled data. We generate fabricated data labeled according to its anomaly class: standard transit, occulted star spot, gravity darkened (GD), and disintegrating rocky exoplanet (DRE). All but the first option is considered an anomaly. Standard transits are simulated using the Mandel and Agol formalism \cite{mandel2002analytic} implemented in the \texttt{batman} \cite{kreidberg2015batman} library, star spot anomalies using the \texttt{ellc} package \cite{maxted2016ellc}, and the GD effects with the \texttt{PyTransit} library \cite{Parviainen2015}. Details about the data generation can be found in Appendix \ref{appendix:data} and physics details in Appendix \ref{appendix:anomalousphysics}. We call our fabricated data the Artificial Labeled Transit dataset (ALT). We produced an idealized version, ALT-i, and a more difficult, or ``hard,'' version ALT-h. The ALT-i version is completely detrended with transits occurring at consistent horizontal shift and scale. The ALT-h version features a vast range of trends and horizontal shifts and scales. Both contain 25,680 training transits with 6,420 standard transits and 6,420 of each anomaly class. Each test sets has 2,700 regular transits and 100 of each anomaly class. The anomalous samples were hand-selected to ensure they are distinguishable from a regular transit. Data examples are shown in Figure \ref{fig:transit_examples}.

\section{Automatic detection of anomalous transit light curves}
\label{sec:method}

We aim to learn a featurization of light curve data which is amenable to unsupervised anomaly detection \cite{yao2019vae_features4anomaly} and incorporates the inductive bias that a transit can be additively decomposed into a \emph{standard transit} component and a \emph{potentially anomalous residual} component. Each piece is modeled by a variational autoencoder (VAE), \emph{TransitVAE} and \emph{ResidualVAE} respectively, such that their sum is a learned reconstruction of the input light curve. 
The latent representation of the ResidualVAE is therefore a clustered and dimensionality-reduced representation of relevant anomalous information.

\paragraph{Method}

The TransitVAE is trained to predict a theoretical transit model according to \cite{mandel2002analytic}. It is always trained on fabricated data to enable using a ground truth general transit shape label $\bxi$. Those labels are always centered, detrended, and noise-free. To better model the observed transit $\bx$ we apply a deterministic transform $f$ with five parameters that accounts for shift, trend and scale of $\bx$. Find implementation details of $f$ in Appendix \ref{appendix:architecture}.

We define TransitVAE's encoder $T^{\enc}_{\bphi_{T}}(\bx) = (\bmu_{T}, \bsigma_T, \bthat)$, where $\bx$ is the light curve data, $\bmu_{T}$ and $\bsigma_T$ are the mean and standard deviation of a normal distribution, $\bthat$ are transformation parameters, and ${\bphi_{T}}$ are the model weights. A posterior predictive sample is produced by taking $\bz_T \sim \mathcal{N}(\bmu_T, \bsigma_T^2 \id)$ and first evaluating the decoder $T^{\dec}_{\btheta_{T}}(\bz_T) = \bxihat$, where $\btheta_{T}$ are model weights and $\bxihat$ is the estimated detrended and centered transit. The estimated observed transit is recovered by $f(\bxihat, \bthat) = \bxhat$. 

ResidualVAE is a traditional VAE. Its latent distribution is defined by encoder $R^{\enc}_{\bphi_{R}}(\by) = (\bmu_{R}, \bsigma_R)$, where $\by = \bx - \bxhat$ is the residual to reconstruct, $\bmu_{R}$ and $\bsigma_R$ parametrize the latent normal distribution and $\bphi_{R}$ are model weights. The reconstruction is estimated by the decoder $R^{\dec}_{\btheta_{R}}(\bz_R) = \byhat$, where $\bz_R \sim \mathcal{N}(\bmu_R, \bsigma_R^2 \id)$ and $\btheta_{R}$ are model weights.

TransitVAE is trained by minimizing the VAE loss where the reconstruction term compares the detrended and centered transit estimate $\bxihat$ to the ground truth $\bxi$. An additional regression loss for the predicted transformation parameters $\bthat$ is applied. ResidualVAE is trained similarly but attempts to reconstruct $\by$ and has no transformation loss. Training occurs in two stages starting by optimizing $\bphi_{T}$ \& $\btheta_{T}$, then TransitVAE is fixed and $\bphi_{R}$ \& $\btheta_{R}$ are optimized. An overview of the losses, the hyperparameter details and a sketch of the design can be found in Appendix \ref{appendix:architecture}.

\section{Experiments on artificial data}

\begin{figure}[h]
  \centering
  \includegraphics[width=\textwidth]{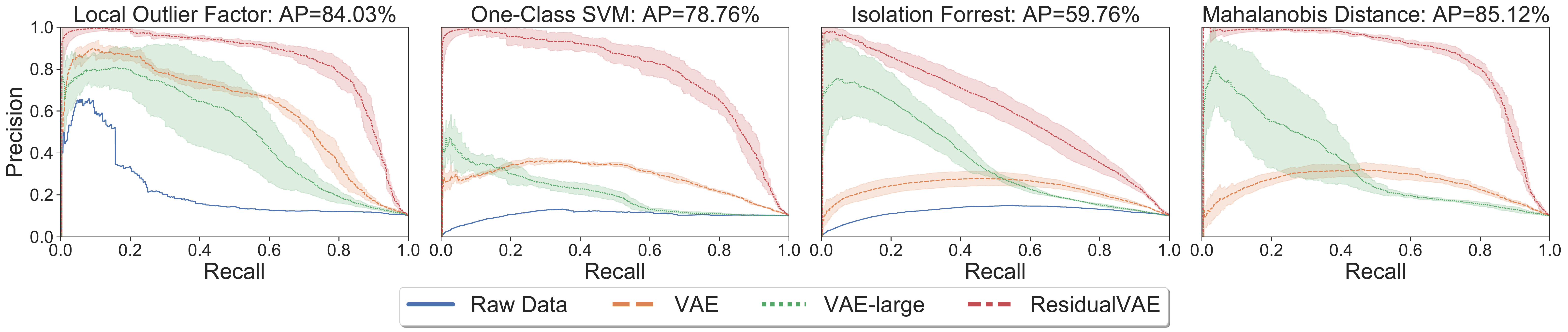}
  \caption{PR curves for various classifiers and featurizations on the ALT-i test set. Mean and 95\% confidence intervals are shown across 5 runs with different random seeds. 
  AP values of the best featurization are given in the title for each classifier. ResidualVAE always had the highest AP.}
  \label{fig:pr_plots}
\end{figure}

\paragraph{Evaluation methods} 
We quantify the performance of each model on artificial, labeled data by plotting the precision versus the recall, i.e. a Precision-Recall (PR) curve. We summarize this with \texttt{scikit-learn}'s formulation of Average Precision (AP), essentially the area under the curve \cite{scikit-learn}.

\paragraph{Setup} We consider the following unsupervised anomaly detection classifiers: Local Outlier Factor (LOF) \cite{breunig2000lof}, the One-Class Support Vector Machine \cite{schoelkopf1999ocsvm}, the Isolation Forrest \cite{liu2008isolation_forrest} and the Mahalanobis distance 
\cite{rousseeuw1999mcd}.
Each classifier is applied to several featurizations including the raw data $\bx$ and the latent representation $\bmu_R$ of ResidualVAE. We trained two more architectures that reconstruct $\bx$ as a benchmark. The first, \emph{VAE}, has as many parameters as ResidualVAE, and the second, \emph{VAE-large}, has the same number of parameters as TransitVAE and ResidualVAE combined. Each classifier is also applied to the mean latent representation of VAE and VAE-large. The procedure was repeated five times with different random seeds to report uncertainty bounds.

\begin{table}[hbt]
    \caption{Average precision (in \%) and standard deviation over 5 independent runs for various classifiers with different featurizations on the ALT-h test set. Each classifier's best result is bolded.}
    \label{tab:average_precision}
    \centering
    \begin{tabular}{lllll}
        \toprule
        {} & Local Outlier Factor & One-class SVM & Isolation forrest & Mahalanobis \\
        \midrule
        Raw data    & 10.59 +/- 0 & 9.29 +/- 0 & 7.83 +/- 0 & - \\
        VAE         & 13.07 +/- 2.12 & 7.78 +/- 0.23 & 8.61 +/- 0.31 & 8.62 +/- 0.33 \\
        VAE-large   & 12.83 +/- 2.04 &  8.30 +/- 0.37 & 9.00 +/- 0.64 &  8.48 +/- 1.32 \\
        ResidualVAE & \textbf{25.10} +/- 3.32 & \textbf{16.37} +/- 1.17 & \textbf{17.20} +/- 2.37 &   \textbf{17.57} +/- 2.91 \\
        \bottomrule
    \end{tabular}
\end{table}

\paragraph{Discussion of results}
ResidualVAE's featurization led to superior precision across nearly all recall levels for every classifier on the ALT-i test set, see Figure \ref{fig:pr_plots}. AP results can be found in Appendix \ref{appendix:additional_results}. 

ResidualVAE outperforms all other methods across all classifiers on the ALT-h test set as seen in Table \ref{tab:average_precision} as measured by AP. ALT-h was more difficult with the highest mean AP reaching only 25\%. This implies that the shifting, scaling, and trending of the data can have a significant effect on performance. The PR curves were qualitatively similar to ALT-i results, thus only AP is reported.

\section{Experiments on TESS data}

\paragraph{Evaluation methods} The Classifier Two-Sample Test (C2ST) \cite{lopezpaz2018c2st, friedman2003multivariate} attempts to distinguish whether two sets of samples are drawn from the same distribution by means of a binary classifier, e.g., a Multi Layer Perceptron (MLP). Assuming the classifier has converged and is sufficiently expressive, two sets of held-out samples drawn from the same distribution will be classified at near-chance level. If the sets are drawn from sufficiently distinct distributions, they will be accurately classified. One can test statistical significance, such as with a p-value, using the accuracy on the held-out data.

\paragraph{Setup}
We apply the TransitVAE that was pre-trained on ALT-h to extract residuals from the real TESS dataset. Then we train the ResidualVAE on the residuals of the real data. We use the LOF scores based on the ResidualVAE features to rank the transits since it performed best on the artificial datasets. To increase robustness of predictions we use an ensemble of five independent runs of TransitVAEs with three independent runs of the ResidualVAE each. 
To assess the quality of the ranking we perform a C2ST comparing various strata to each other. Our hypothesis is that highly-ranked strata are drawn from a different distribution (outliers) than lower-ranked strata. For example, an equally sized set of held-out samples from the top 10\% (90th percentile) are compared to the bottom 90\%. A high C2ST with a small p-value provides some evidence that the 90th percentile contains outliers. As classifier we use a simple MLP with two hidden layers (64 and 10 neurons). As input we use the raw transit data and evaluation is done with five-fold cross validation. Our significance threshold is set to $p \leq 0.05$.

\begin{table}[hbt]
    \caption{C2ST accuracies and p-values for various strata of real TESS transits with ResidualVAE featurizations. Samples from the identified subset are compared to samples from the remainder of the data. Rank is determined by LOF score. Statistically significant results, $p \leq 0.05$, are bolded.
    }
    \label{tab:c2st}
    \centering
    \begin{tabular}{lllll}
        \toprule
        Percentiles & Up to 90th & Up to 50th & 80th through 65th & 65th through 50th \\
        \midrule
        Accuracy (in \%)    & \textbf{71.14} & \textbf{65.77} & \textbf{64.00} & 56.80 \\
        p-value         & $1.44e^{-4}$ & $2.34e^{-9}$ & $1.76e^{-3}$ & $9.67e^{-2}$ \\
        \bottomrule
    \end{tabular}
\end{table}

\paragraph{Discussion of results}
Table \ref{tab:c2st} shows the classifier accuracies and corresponding p-values comparing samples from the identified subset to the remainder of the data. Tests sampling strata from the top and middle of the ranking are shown to confirm that the highly ranked data contains more anomalies than the middle rankings. All tests resulted in a classification accuracy above chance and all but one was considered significant by our criterion. 

\section{Conclusion}
\label{sec:conclusion}
In this paper we introduced the first automated approach for anomalous transit shape detection and evaluated it on artificial and real data. On artificial data, we showed that ResidualVAE extracted features better suited to unsupervised anomaly detection methods than using raw data or featurizations from a standard VAE. ResidualVAE identified statistically significant outliers in real data.

\paragraph{Limitations} Our method depends on estimates of epoch, period and transit duration to extract the transits from light curve data. Current estimates provided by the space telescope's processing pipeline may suffer from inaccuracies which leads to shifts and scale mismatches. While the deterministic transform $f(\bxihat, \bthat)$ is designed to mitigate this effect, its success is limited by the accuracy of the predictions for $\bthat$. Errors in $\bthat$ cause artifacts in the residuals which can negatively affect the performance. We do not present any results on TransitVAE's performance of predicting $\bt$ because without any reference for comparison the regression loss is less meaningful. \\ 
Despite the efforts made in creating realistic fabricated data it is always possible that the real data distribution does not align with our simulated one. This might introduce unwanted biases. 

\paragraph{Future Work}
A direction for future work could be improving the accuracy of predicting $\bt$, since correctly aligning the theoretical transit shape with the observed data before calculating the residuals is essential for the performance of our method. 
Other works on anomaly detection use the reconstruction error of an AE or the reconstruction probability of a VAE
as metric for separating anomalies \citep{an2015rec_anomaly}. This would enable independence from traditional anomaly classifiers. We did not pursue this direction in our work but experimenting with such an approach and comparing the results could be done in future work.

\paragraph{Social Impact}
Our method uses existing telescope data and does not have application outside of exoplanetary science. The main positive impacts would be reducing the effort of astronomers for visual inspection of data and potentially inspiring future scientists. The main negative impact is the (relatively small) cost to the environment of training on GPUs.

\section*{Acknowledgements}
This paper includes data collected with the TESS mission, obtained from the MAST data archive at the Space Telescope Science Institute (STScI). Funding for the TESS mission is provided by the NASA Explorer Program. STScI is operated by the Association of Universities for Research in Astronomy, Inc., under NASA contract NAS 5–26555.

\small
\bibliographystyle{unsrt}
\bibliography{bibliography}

\clearpage
\appendix
\section{Data details}
\label{appendix:data}

\subsection{Genuine data}
We use the targets from the TESS Objects of Interest (TOI) catalogue. It consists of known exoplanets identified by previous missions and planetary candidates detected by TESS.
At the time this work was conducted the TOI catalogue contained about 2600 targets. Using the \textit{astroquery} package \cite{ginsburg2019astroquery} we were able to retrieve data for 1668 of those targets by querying their TESS Input Catalogue (TIC) IDs in MAST \cite{mast}. We use the PDCSAP flux channel generated by the TESS pipeline which contains some data correction.
Not a Number (NaN) values and statistical outliers (values higher than five standard deviations from the mean flux) are removed from the light curve. 
We phase-fold the light curve according to the estimated orbital period and extract the cadences within a time window of three transit durations centered around the transit midpoint. This is because some anomalous features might occur shortly before or after the transit and out-of-transit data can give some context about possible trends in the data. The parameter estimates for the period, epoch and transit duration are obtained from the respective data validation file if available or directly from the TOI table in NASA's exoplanet archive \cite{nasa2021exoplanet_archive} otherwise. The folded light curve segments are binned to 256 equally spaced cadences and data gaps are filled with linear interpolation if present. The values of each transit shape are scaled to be in a range between -1 and 1.

\subsection{Fabricated data}
Realistic transit data can be synthesized via theoretical transit models based on mathematical formalisms such as the one of Mandel and Agol \cite{mandel2002analytic}. Fabricated transits have the same format as the real data (vector size of 256 and range of values between -1 and 1).

\paragraph{Regular transits}
We use the \texttt{batman} library by Kreidberg \cite{kreidberg2015batman} which requires the following parameters:
planet-star radii ratio $r$, semi-major axis in stellar radii $a$, orbital inclination angle $i$, orbital eccentricity $e$, longitude of periastron $\bar{\omega}$ and the limb darkening coefficients. For our simulations we use the quadratic limb darkening law which has two coefficients $c_{1}$ and $c_{2}$.
Limb darkening coefficients are obtained via a lookup table for TESS photometry provided by \cite{claret2017limb_TESS}. This requires knowledge about the stellar surface temperature $T$, surface gravity $\log g$ and metalicity $Z$.
A uniform distribution over those parameters might not reflect the general transit shapes observed in real data well. Hence, we sample constellations of confirmed exoplanets from a table with parameter estimates available at NASA's exoplanet archive \cite{nasa2021exoplanet_archive}. For some systems certain parameter estimates are missing. In these cases we uniformly sample them from a plausible range of values. An overview of parameter ranges for missing values is presented in Table \ref{tab:regular_params}.
\begin{table}[htb!]
    \caption{Parameter ranges for missing planetary system parameters during artificial transit synthesis.}
    \label{tab:regular_params}
    \centering
    \begin{tabular}{lllllll}
        \toprule
        {} & $R^*$ & $R_{p}$ & $a$ & $b$ & $e$ & $\bar{\omega}$ \\
        \midrule
        Minimum & 0.3 & 0.3 & 2.3 & 0.0 & 0.0 & 0.0 \\
        Maximum & 4.0 & 7.0 & 30.0 & 1.0 & 1.0 & 90.0 \\
        Unit    & $R_{\odot}$ & $R_{e}$ & $R^*$ & $R^*$ & - & degrees\\
        \bottomrule
    \end{tabular}
\end{table}
The radius ratio is obtained by $r=\frac{R_{p}}{R^*}$, where $R_{p}$ is the planetary radius and $R^*$ is the stellar radius. The inclination is obtained by $i=\arccos(\frac{b}{a}) \frac{180}{\pi}$, where $b$ is the impact parameter.
$T$, $\log g$ and $Z$ are determined sequentially in this order. The range of values for each of those parameters is constrained by the combination of values available in \cite{claret2017limb_TESS}.

The theoretical transit models do not take into account photon noise or other sources of disturbances. To make the data more similar to real data we add Gaussian-distributed noise to every data point. For each modeled transit we uniformly sample a signal-to-noise ratio (SNR) in the range between 1 and 30. With SNR we refer to the ratio between the maximal transit depth of the signal divided by the standard deviation of the added Gaussian noise. The range of SNRs is inspired by the values observed in TESS data. Note that in TESS data the distribution over different SNRs is not uniform. 
There can occur trends in the data that are due to stellar variability (e.g due to rotational modulation through star spots). To simulate this effect we add a linear trend to half of the generated signals. The slope of the trend can be positive or negative and its magnitude is maximally two times the transit depth within the observation window. The generation of anomalous transits is analogous, just using appropriate software designed for the anomaly class. 

\paragraph{Transits with (occulted) star spots}
The software package \texttt{ellc} \citep{maxted2016ellc} supports the modelling of circular star spots. It requires further parameters for each spot: the longitude $\lambda_{s}$ and latitude $\phi_{s}$ on the stellar surface, the radius $R_{s}$ of the spot, the brightness factor $b_{s}$ of the spot relative to the stellar surface and the rotation period $p^*$ of the star.

Our transit models with star spot features contain up to four star spots with a higher probability for a lower number of star spots. Configurations with more than one spot are always half as likely as with one spot less. $\phi_{s}$ is biased to lie close to the latitude of the planet's transit which is parametrized by the impact parameter $b$. Their latitude has an offset from the line of transit no larger than five degrees in either direction. This ensures that the transit models contain examples of occulted star spots but there will be also configurations where no spot is occulted and only the modulation by the stellar rotation is visible. In the test set, only examples that exhibit occulted spot features are selected.

$\lambda_{s}$ is uniformly sampled in a range between -60 and 60 degrees from the center of the stellar disk. This is because further at the limb of the star the effect of occulted star spots can become extremely small due to interaction with the limb darkening effect. We uniformly sample $b_{s}$ in the range between 0.7 and 1.3 modeling both dark and bright spots with a realistic magnitude. The sizes are sampled dependent on each other to reduce the probability of overlapping star spots and avoid too much of the surface being covered by star spots. The maximal spot radius size for the first spot is 20 degrees, which is reduced by five degrees for every further spot. Each spot radius is sampled uniformly in a range between two degrees and the individual maximal radius.
$p^*$ is taken from the list of parameters if available or uniformly sampled from a range between 10 and 40 days otherwise. 

\paragraph{GD transits} For modelling the gravity darkening effect on the transit shape we use \texttt{PyTransit} which offers an oblate star model. Additional parameters for this model are the gravity darkening coefficient $\beta$, the temperature at the poles $T_{pole}$, the stellar inclination $i^{*}$ and the sky-projected obliquity $\lambda$.
We obtain $\beta$ from \cite{claret2017limb_TESS} analogously to the limb darkening coefficients. We uniformly sample the pole temperature from a range between 5700 and 12 000 Kelvin. The parameters $i^{*}$ and $\lambda$ are uniformly sampled from a range between zero and 90 degrees which covers all possible orientations that would result in a different effect (the remaining constellations are redundant due to symmetry).

Gravity darkening effects reported in the literature have usually magnitudes between 1e-3 and 1e-4 in a normalized light curve.
For each configuration of $i^{*}$, $\lambda$ and $\beta$ we test up to ten different values for $T_{pole}$. If a configuration shows an effect larger than 1e-3 or smaller than 1e-4 it is skipped. After ten unsuccessful configurations the current stellar parameters are discarded and the next parameter set is tested. 

\paragraph{DRE transits} Accurate theoretical modeling of DREs is not as well understood as the other phenomena. This is why we decided to use real world examples to model DREs. Specifically, we use light curves of Kepler-1520 b, a known DRE, to create residual profiles from its transits and the best-fit theoretical model from the Kepler pipeline. Those residual profiles are added to regular theoretical transit models and the scale of the SNR is sampled in the same way as for regular transits. To increase the variety in DRE samples we add a small Gaussian noise with standard deviation $\sigma < 0.5 res$, where $res$ is the average magnitude of the residual profiles. This ensures diverse examples while not overshadowing the internal structure of the residual profile.

\section{Physics of anomalous transits}
\label{appendix:anomalousphysics}
To give readers interested in the astronomical part of this work a better understanding of the anomaly classes defined for our artificial datasets, we shortly elaborate on the astrophysical source of the anomalous signal and possible indications for astronomy research. For regular transits the largest variability in shape comes from the \textit{limb-darkening} effect and the \textit{impact parameter} $b = \frac{a*\cos(i)}{R_{*}}$, where $a$ is the semi-major axis of the system, $i$ is the orbital inclination angle with respect to our point-of-view and $R_{*}$ is the stellar radius. The limb-darkening effect causes the star to appear brighter at the center and dimmer at the edges (or limb). An impact parameter of $b=0$ refers to a planet that transits along the equator of the star producing a U-shaped transit while a high impact parameter corresponds to a grazing planet that transits at the edge of the stellar disk and produces a shallower more V-shaped transit. The difference in shape is caused by the different ratios between ingress/egress (partial occlusion of planet at beginning/end of transit) duration and full occlusion duration.
In the case of occulted star spots dark or bright areas on the stellar surface can temporarily change the amount of light that is blocked by the transiting planet resulting in "bumps" in the transit shape. Accurate modeling of occulted star spots can for example increase the precision of radius estimates for the planet. Gravity darkening happens when a star is rapidly rotating around its own axis. The centrifugal forces drag the poles closer to the center and push away mass at the equator making the star more ellipsoid shaped. This causes hotter, brighter poles and a cooler, dimmer "belt" at the equator. Depending on the angle of the planetary orbit with respect to the rotation axis of the star from our point of view (projected obliquity) this can cause asymmetries in the transit shape. Based on those asymmetries the projected obliquity can be determined. DREs drag a gas/dust tail behind them that blocks additional light. An exponential decay of the light blocked by the tail leads to a characteristic asymmetry of ingress and egress as well as a bump shortly before ingress that is produced by light reflected from particles. Follow-up studies of DREs can answer questions about their internal composition. 

\section{Architecture details and hyperparameters}
\label{appendix:architecture}

\subsection{Architecture, hyperparameters and compute}
\begin{figure}[hb!]
  \centering
  \includegraphics[width=0.8\textwidth]{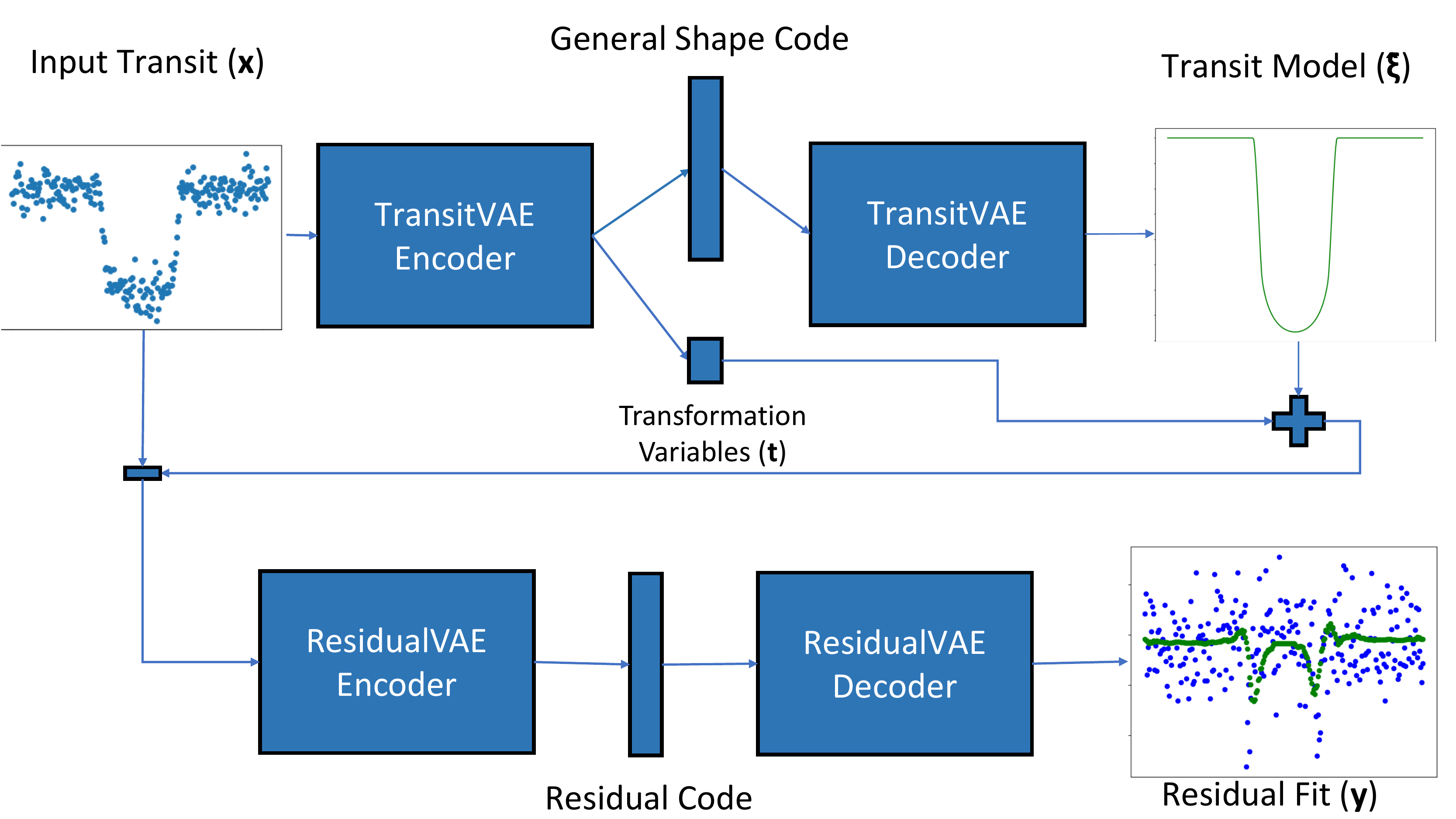}
  \caption{Step-wise feature extraction by TransitVAE and ResidualVAE. "+" represents the application of the deterministic transform $f$ (see eq. \ref{eq:transform}) and "-" represents the calculation of the residuals.}
  \label{fig:two_step_model}
\end{figure}

\begin{figure}[ht!]
  \centering
  \includegraphics[width=0.8\textwidth]{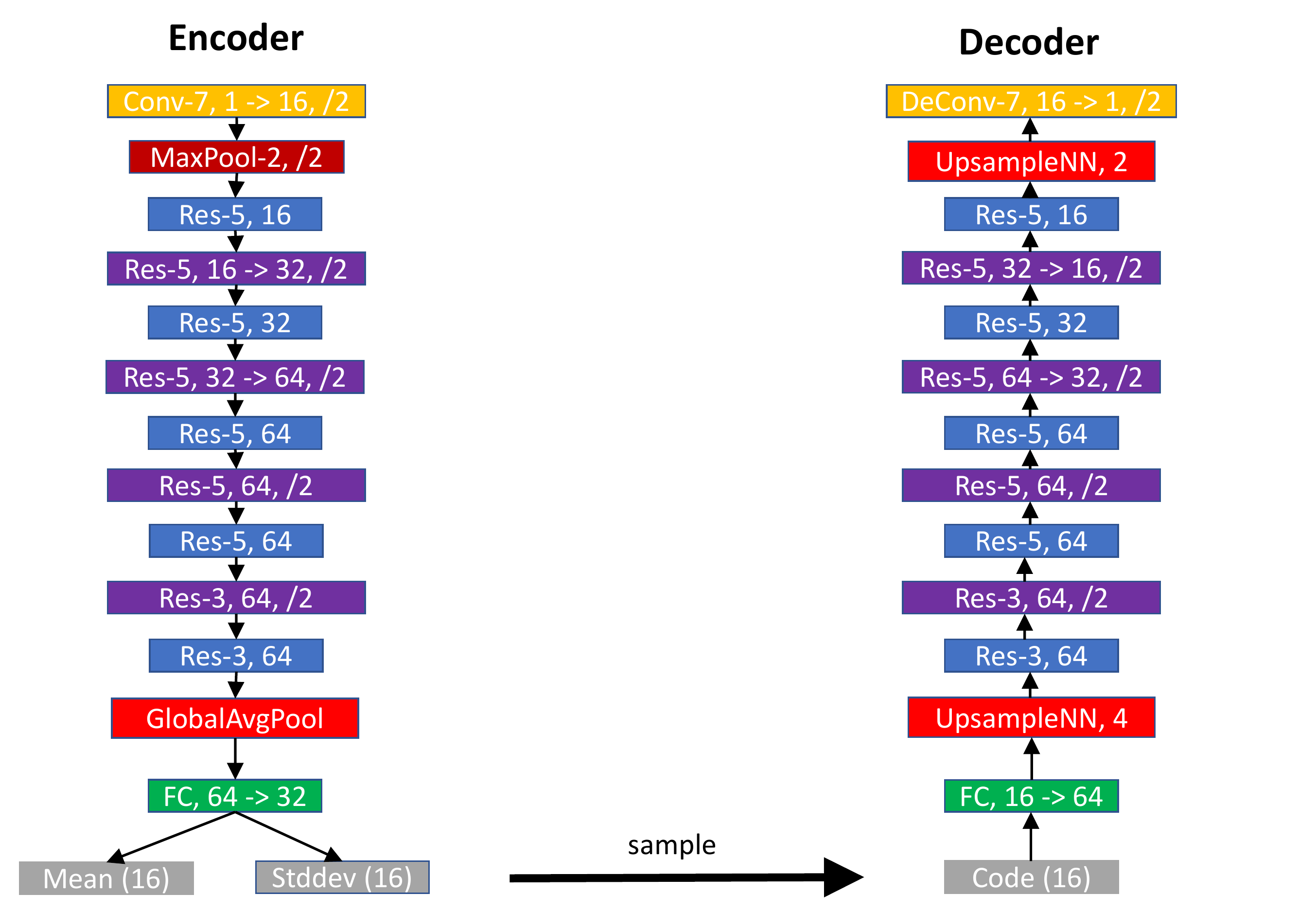}
  \caption{VAE Architecture. Conv refers to a 1D Concolutional layer, DeConv to a transposed 1D convolution. The first integer describes the kernel size and the second the number of filters. 1->16 refers to one input and 16 output channels, /2 refers to a stride of 2. All layers are followed by a batch normalization layer and a ReLU activation function. Res describes a Residual block with two Conv layers in the encoder and DeConv layers in the decoder. FC ia a fully connected layer}
  \label{fig:architecture}
\end{figure}

The illustration of our approach is summarized in Figure \ref{fig:two_step_model}. The detailed neural architecture of the VAE is summarized in Figure \ref{fig:architecture}.

This architecture has about 527k trainable parameters. It is used for the TransitVAE, ResidualVAE and standard VAE. Note that the Transit VAE has a few more trainable parameters due to the additional output of the five transformation parameters, which is negligible. For the VAE-large the number of filters is changed from 16, 32 and 64 to 23, 46 and 92 respectively to yield a total of about 1.09M trainable parameters which is roughly the number of parameters of the TransitVAE and ResidualVAE combined.

The models are implemented in \texttt{PyTorch} \cite{NEURIPS2019_9015PyTorch} (version 1.7.1). We use the Adam optimizer with an initial learning rate of $5e^{-4}$ and \texttt{PyTorch}'s default parameters for the weight updates. Training was executed on a single GPU (2GB RAM) with CUDA 10.0 of a modern notebook. This took about 40 minutes for 100 epochs of one of the ALT training sets. VAE-large was trained for 200 epochs which took about 80 minutes. We make use of a dynamical weighting between the reconstruction loss and the Kullback-Leibler divergence loss as described in \cite{vae_gamma_weight}.

\subsection{Details on deterministic transform $f$}

The deterministic transform is defined as:
\begin{align}
    \label{eq:transform}
    f(\bxi,\bt) &= \trend(\scale_v(\shift_h(\scale_h(\bxi, s_{h})
    ,l_{h}),s_v),t_{s},t_{e}) \: \text{with} \: (s_{h},l_{h},s_v,t_{s},t_{e})^{T} = \bt.
\end{align}

Recall $\bxi$ is the predicted general transit (noise-free, centered and detrended) and $\bt$ are the transformation parameters consisting of the horizontal scale $s_{h}$, the horizontal shift (or lag) $l_{h}$, the vertical scale $s_{v}$, the start value of the trend $t_{s}$ and end of value of the trend $t_{e}$.
First the horizontal scale $s_{h}$ is applied according to $\scale_h(\bxi, s_{h}) \equiv \cut(\resize(\bxi, s_{h}n))$, where $\resize$ corresponds to resizing with linear interpolation, $n$ is the size of $\bxi$ and $\cut$ applies either center-cropping to size $n$ if $s_{h}>=1$ or applies edge padding equally to each side up to size $n$ if $s_{h}<1$. 
The horizontal shift is applied via cyclic permutation $\shift_h(\bxi,l_h) \equiv \text{for each} \: \xi_{i} \in \bxi : i = \mod(i+l_{h}, n)$, where $\mod()$ is the modulo operation, $i$ is an index of $\bxi$ and $n$ is its size.
The vertical scale is applied by $\scale_v(\bxi,s_v) \equiv s_{v}(\bxi-\max(\bxi))+\max(\bxi_{h})$. A linear trend is applied by $\trend(\bxi,t_s,t_e) \equiv \bxi + \interpl(t_s, t_e, n)$, where $\interpl$ is a linear interpolation between the first two arguments of size $n$, which is the size of $\bxi$.
Since the $\scale_h$ and $\shift_h$ operations are non-differentiable $s_{h}$ and $l_{h}$ are learned via regression loss on their ground truth values (only available for fabricated data). Parameters $s_v$, $t_s$ and $t_e$ are learned via mean-squared error loss between $\bxi$ and input data $\bx$.

\subsection{Overview of Losses}
Transformation Loss:
\begin{equation*}
    L_{\text{transform}} = 2(MSE(\hat{l_{h}}, l_{h}) + MSE(\hat{s_{h}}, s_{h})) + MSE(\bxhat, \bx)
\end{equation*}
TransitVAE Loss:
\begin{equation*}
    \frac{MSE(\bxihat, \bxi)}{2\gamma^{2}} + \beta KL(\bmu_{T}, \bsigma_{T}) + \lambda L_{\text{transform}}
\end{equation*}
ResidualVAE Loss:
\begin{equation*}
    \frac{MSE(\byhat, \by)}{2\gamma^{2}} + \beta KL(\bmu_{R}, \bsigma_{R})
\end{equation*}
For an explanation of the variables see section \ref{sec:method} and the preceeding section about the deterministic transform $f$. $MSE$ is the mean squared error and $KL(\cdot)$ refers to the analytical form of the Kullback-Leibler Divergence between a Gaussian with diagonal covariance matrix and a standard normal distribution. The dynamical recunstruction term weight $\gamma$ is a monotonically decreasing estimate of the current reconstruction loss as defined in \citep{vae_gamma_weight}. $\beta$ is a fixed weight of the KL term that was tuned for each model. For the VAE-large model we used $\beta=1$, for all others $\beta=0.1$. $\lambda$ is a fixed loss weight for the transformation loss, which should be high in order to achieve quick learning of the semantics of shift, scale and trend. The performance of the models is not very sensitive towards this parameter, we chose $\lambda=64$. \\

\subsection{Anomaly detection details}
For the unsupervised anomaly detection methods we use \texttt{scikit-learn}'s \citep{scikit-learn} implementation of the classifiers with default settings. For the LOF we follow the recommendation from the original paper to determine the final score of each example by taking the maximum over several evaluations of the LOF with different values for $k$, where $k$ defines the number of nearest neighbours that are used to estimate the score. We evaluate the LOF for $k \in \{15, 20, 25, 35, 50\}$ which are samples within the range recommended by the authors.

\section{Additional results}
\label{appendix:additional_results}
Table \ref{tab:ap_alt_i} shows the full AP results on the ALT-i dataset for each classifier on different featurizations. Reported values correspond to the mean and standard deviation over five independent runs with different random seeds. The ResidualVAE features perform the best across all classifiers. The Mahalanobis distance with ResidualVAE features achieves the overall best score closely followed by the LOF with ResidualVAE features.

\begin{table}[htb!]
    \caption{Average precision (in \%) and standard deviation over 5 independent runs for various classifiers with different featurizations on the ALT-i test set. Each classifier's best result is bolded.}
    \label{tab:ap_alt_i}
    \centering
    \resizebox{\textwidth}{!}{
        \begin{tabular}{lllll}
            \toprule
            {} & Local Outlier Factor & One-class SVM & Isolation forrest &     Mahalanobis distance \\
            \midrule
            Raw data    & 22.22 +/- 0 & 10.40 +/- 0 & 12.01 +/- 0 & - \\
            VAE         & 60.95 +/- 1.29 &  28.35 +/- 0.60 & 22.19 +/- 2.41 &  25.18 +/- 2.43 \\
            VAE-large   & 50.04 +/- 10.85 &  20.73 +/- 1.86 & 37.88 +/- 3.87 &  34.15 +/- 8.44 \\
            ResidualVAE & \textbf{84.03} +/- 1.97 & \textbf{78.76} +/- 4.29 &  \textbf{59.76} +/- 4.23 & \textbf{85.12} +/- 1.41 \\
        \bottomrule
        \end{tabular}
    }
\end{table}

\end{document}